\documentclass[conference]{IEEEtran}
\IEEEoverridecommandlockouts
\usepackage{cite}
\usepackage{amsmath,amssymb,amsfonts}
\usepackage{algorithmic}
\usepackage{graphicx}
\usepackage{booktabs}
\usepackage{url}
\usepackage{multirow}
\usepackage{makecell}
\usepackage{textcomp}
\usepackage{xcolor}
\usepackage{bbm}
\usepackage{caption}
\usepackage{cleveref}
\usepackage[left=0.66in,right=0.66in,top=0.75in,bottom=1.04in]{geometry}
\def\BibTeX{{\rm B\kern-.05em{\sc i\kern-.025em b}\kern-.08em
    T\kern-.1667em\lower.7ex\hbox{E}\kern-.125emX}}
\begin{document}

\title{Exploring Semantic Clustering and Similarity Search for Heterogeneous \mbox{Traffic Scenario Graphs}}

\makeatletter
\def\ps@IEEEtitlepagestyle{%
  \def\@oddfoot{\mycopyrightnotice}%
  \def\@evenfoot{}%
}
\def\mycopyrightnotice{%
  \begin{minipage}{\textwidth}
  \centering \scriptsize
  Copyright~\copyright~2025 IEEE. Personal use of this material is permitted. Permission from IEEE must be obtained for all other uses, in any current or future media, including reprinting/republishing this material for advertising or promotional purposes, creating new collective works, for resale or redistribution to servers or lists, or reuse of any copyrighted component of this work in other works.
  \end{minipage}
}
\makeatother

\author{
Ferdinand~Mütsch$^{1}$,
Maximilian~Zipfl$^{2}$,
Nikolai~Polley$^{1}$,
and J.~Marius~Zöllner$^{1,2}$

\thanks{$^{1}$KIT Karlsruhe Institute of Technology, Karlsruhe, Germany
{\tt\small \{muetsch, polley\}@kit.edu}}%
\thanks{$^{2}$FZI Research Center for Information Technology, Karlsruhe, Germany
{\tt\small \{zipfl, zoellner\}@fzi.de}}%
}

\maketitle

\begin{abstract}
Scenario-based testing is an indispensable instrument for the comprehensive validation and verification of automated vehicles (AVs). However, finding a manageable and finite, yet representative subset of scenarios in a scalable, possibly unsupervised manner is notoriously challenging. Our work is meant to constitute a cornerstone to facilitate sample-efficient testing, while still capturing the diversity of relevant operational design domains (ODDs) and accounting for the "long tail" phenomenon in particular. To this end, we first propose an expressive and flexible heterogeneous, spatio-temporal graph model for representing traffic scenarios. Leveraging recent advances of graph neural networks (GNNs), we then propose a self-supervised method to learn a universal embedding space for scenario graphs that enables clustering and similarity search. In particular, we implement contrastive learning alongside a bootstrapping-based approach and evaluate their suitability for partitioning the scenario space. Experiments on the nuPlan dataset confirm the model's ability to capture semantics and thus group related scenarios in a meaningful way despite the absence of discrete class labels. Different scenario types materialize as distinct clusters. Our results demonstrate how variable-length traffic scenarios can be condensed into single vector representations that enable nearest-neighbor retrieval of representative candidates for distinct scenario categories. Notably, this is achieved without manual labeling or bias towards an explicit objective such as criticality. Ultimately, our approach can serve as a basis for scalable selection of scenarios to further enhance the efficiency and robustness of testing AVs in simulation.
\end{abstract}


\section{Introduction}

Scenario-based testing in simulation is crucial during development and validation of highly automated or autonomous vehicles (AVs). Its purpose is to gain confidence in correct behavior and to provoke failures \cite{hauer_did_2019}. However, since the space of all conceivably possible scenarios is infinitely large, a key question revolves around which particular scenarios to include as test cases to combat this open-context problem. The well-known \textit{long tail} phenomenon poses an additional challenge in that particularly interesting corner-case scenarios are especially rare to find in real-world data. Traditionally, experts relied on hand-crafted scenario catalogs. However, their substantial limitations in scalability and diversity motivate the need to go beyond pure expert knowledge and address the problem in a data-driven fashion instead, using rule-based, machine learning-based (ML), or hybrid approaches~\cite{weber_toward_2023}. Besides pure extraction, it is important to group scenarios into distinct classes and thus reduce the scenario space down to a manageable, finite, yet representative test set. Sampling from these clusters enables for more efficient testing, whereby \textit{``the more complete a set [...] is, the more convincingly testing can ensure correct system behavior''} \cite{hauer_clustering_2020}. A prerequisite for effective clustering are meaningful representations of scenarios which support comparison by similarity. Notably, this similarity is supposed to particularly capture scenarios' \textit{semantics}, presuming that even though two scenarios differ \textit{syntactically}, they may nonetheless depict the same real-world situation. Moreover, the notion of similarity that we strive for is, ideally, not restricted to a previously chosen objective criterion such as \textit{criticality}, but can rather reflect also more nuanced, latent and less obvious correlations that a human might not identify~\cite{hauer_did_2019}.

\begin{figure}
    \includegraphics[width=\linewidth]{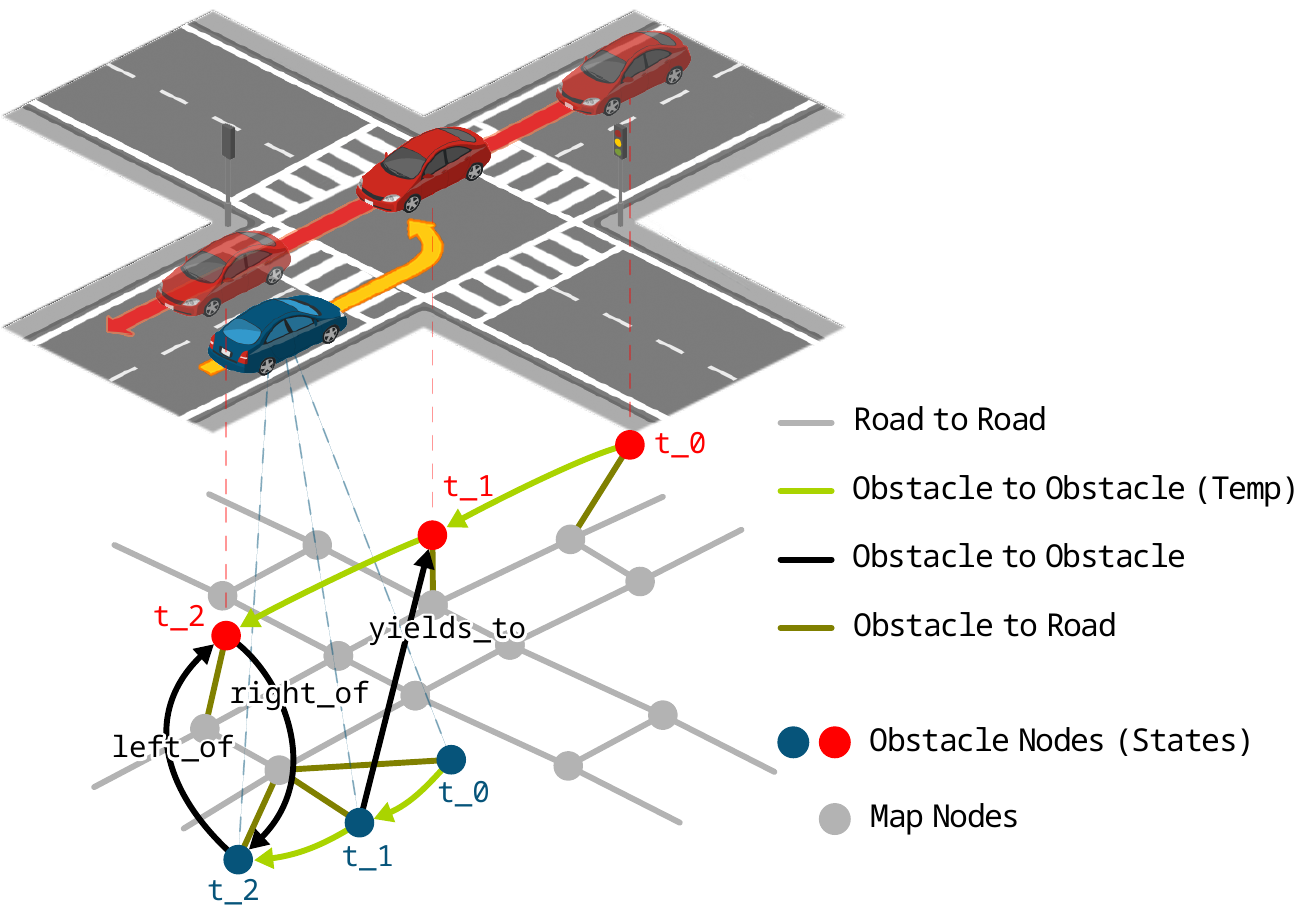}
    \centering
    \caption[Heterogeneous scenario graph schema]{Heterogeneous scenario graph schema}
    \label{fig:scenario_graph}
    \vspace{-0.4cm}
\end{figure}

Existing approaches to extraction and clustering are mostly rule-based and either restricted to very few ODDs (e.g. highways), only consider a fraction of scenarios' feature information (e.g. only basic positional information) or necessitate ``hard-coded'' comparison criteria. Building upon our graph-based data model, we contribute a method to learn descriptive latent representations of variable-length scenarios for similarity search and showcase their suitability for clustering without an explicit objective function. For this purpose, we also extend the self-supervised learning method presented by Thakoor et al. \cite{thakoor_large-scale_2023} to make it work for entire (heterogeneous) graphs. Eventually, all of this enables to compile a comprehensive, extensible scenario database in an automated fashion. 

\section{Related Work}

\subsection{Scenario Representation}
\label{sec:2.1}
Literature has different modalities to model and describe traffic scenarios. Besides high-level formal descriptions like OpenSCENARIO\footnote{\url{https://www.asam.net/standards/detail/openscenario-dsl/}} or abstract time series and object lists \cite{weber_toward_2023, hauer_clustering_2020}, scenarios are often represented in image space \cite{li_bevformer_2025, kim_drivegan_2021} or as graphs. In this work, we opt for a graphical representation with the benefit of such being particularly flexible to encode properties on nearly all layers of a scenario, including relational aspects like interactions among actors. Excellent prior work has already been carried out in this regard. The graph-based data models proposed as CRGeo \cite{meyer_geometric_2023}, HDGT \cite{jia_hdgt_2023} and HoliGraph \cite{grimm_heterogeneous_2023} are conceptually similar and mainly differ in the way they encode spatial- and temporal information. Depending on the targeted downstream tasks, key design choices include:

\begin{itemize}
    \item \textbf{Map embedding:} Map information can either be embedded in the graph itself (all above works) or referenced in separate layer. Optionally, traffic light states and traffic rules may be added.
    \item \textbf{Spatial coordinates:} Spatial positions can be encoded as absolute coordinates, relative to some common reference \cite{meyer_geometric_2023} or pair-wise relative \cite{jia_hdgt_2023, grimm_heterogeneous_2023}. This also dictates whether scenarios are scene- or agent-centric.
    \item \textbf{Temporal aspects:} For encoding scenario dynamics, one option is to employ a dynamic graph by essentially ``stacking'' multiple \textit{scene} graphs along the temporal dimension. Alternatively, in \textit{temporal unrolling} \cite{wu_graph_2022}, agent nodes are replicated for each time step and interconnected via temporal edges \cite{grimm_heterogeneous_2023, meyer_geometric_2023}, as depicted in \Cref{fig:scenario_graph}. As a third option, \cite{jia_hdgt_2023} encodes agent history (or future) using a fixed-length latent per-node feature vector. While the latter is very explicit and compact it is not as flexible as to change at runtime, however.
\end{itemize}

A minor limitation of HoliGraph in view of scenario similarity learning is its agent-centric nature and the necessity of map information. CRGeo, in turn, only supports fixed-length scenarios in its current state. ``Semantic Scene Graph'' \cite{zipfl_towards_2022} and roadscene2vec \cite{malawade_roadscene2vec_2022} are restricted to single scenes only.

\subsection{Scenario Clustering}
As a prerequisite for efficient AV testing, the problem to cluster traffic scenarios into cohesive groups by semantic similarity has been approached multiple times already. Weber et al. \cite{weber_toward_2023}  implement a five-step process for this. In a spatio-temporal filtering step, they first identify relevant subsequences of long-running scenarios based on threshold values for (a) pair-wise Euclidean distances between actors and (b) scene criticality (using post-encroachment time). During subsequent feature extraction, scene-wise occupancy grids (centered around the ego vehicle) are constructed, fed through a principal-component analysis (PCA) for dimensionality reduction and stacked up to 4-dimensional tensors across space and time. Eventually, agglomerative hierarchical clustering is applied on these. VistaScenario \cite{chang_vistascenario_2024} places a particular focus on the extraction step and proposes an elaborate way to segment recordings into ``atomic'' scenarios using their spatio-temporal evolution tree, followed by subsequent clustering. Similarly, Hauer et al. \cite{hauer_clustering_2020} apply dynamic time-warping (DTW) to prepare agent trajectories for the subsequent k-Means clustering, which is eventually evaluated with respect to its silhouette score. Kruber et al. \cite{kruber_unsupervised_2018} employ a random forest in conjunction with hierarchical clustering on simulated critical scenarios that comprise two time steps and two obstacles each, represented by ten basic pair-wise features.

All aforementioned approaches, however, use hand-picked features or only exploit very basic occupancy- or trajectory information. In contrast, we seek to retain as much information from the scenario as possible -- including relational aspects – and therefore choose to learn feature relevance from data, as opposed to manual feature engineering. This is conceptually similar to Zipfl et al. \cite{zipfl_traffic_2023}, who employ self-supervised, contrastive learning for latent feature extraction, followed by hierarchical clustering of the embedding space to obtain groups of individual scenes. Their analyses hint at the effectiveness of akin approaches by revealing a correlation between principal components of the embeddings and ground-truth scene graph attributes as well as a correspondance between clusters and human-interpretable classes of scenes. However, their work is limited to individual scenes, while the purpose of this work requires to consider entire scenarios.

\subsection{Graph Similarity (Learning)}
Since we represent scenarios as graphs, the problem of measuring similarity between scenarios can be approached with general graph similarity learning (GSL) techniques, which Li et al. \cite{li_graph_2019} give a good primer on. A simple similarity measure is graph edit distance (GED), but, in its original form, is only applicable to small, non-attributed graphs. Other common methods usually involve to first compute a kernel embedding and then apply some well-known distance metric (e.g. L2-norm) to obtain the kernel value, thus referred to as vector-based kernel methods. Goal is to find fixed-sized, low-dimensional vector representations such that distances in the vector space reflect semantic similarity with similar samples located close together and dissimilar ones residing far apart. Traditional kernels (e.g. Weisfeiler-Lehman subtree or random walks) as well as graph2vec \cite{narayanan_graph2vec_2017} operate solely on the graphs' structure, rendering them impractical for our use case. In the realm of deep representation learning, generative methods, including (permutation-invariant) graph autoencoders (e.g. PIGVAE \cite{winter_permutation-invariant_2021}), and self-supervised discriminative methods -- often times following the contrastive learning paradigm (e.g. GraphCL \cite{you_graph_2020}, GRACE \cite{zhu_deep_2020}) -- have gained popularity. To bypass the struggle of choosing appropriate negative samples for contrastive methods, BGRL \cite{thakoor_large-scale_2023} instead applies the \textit{bootstrapping} idea to graphs. Lastly, graph matching networks (GMNs) have emerged as a promising way to approximate the traditionally very expensive graph matching process (also required for GED) with GNNs \cite{li_graph_2019}.

In view of the size and structural complexity of traffic scenario graphs, the aforementioned traditional approaches are not suitable for the problem at hand. Unsupervised, learning-based approaches, on the other hand, show promising results. Therefore, we choose BGRL as a basis for our model and additionally implement another variant based on GraphCL for comparison -- primarily to better understand the effectiveness of not having to construct negative samples.
\section{Concept}
\label{sec:3}

\subsection{Graph Data Model}
\label{sec:3.1}

\begin{figure*}
    \includegraphics[height=56mm]{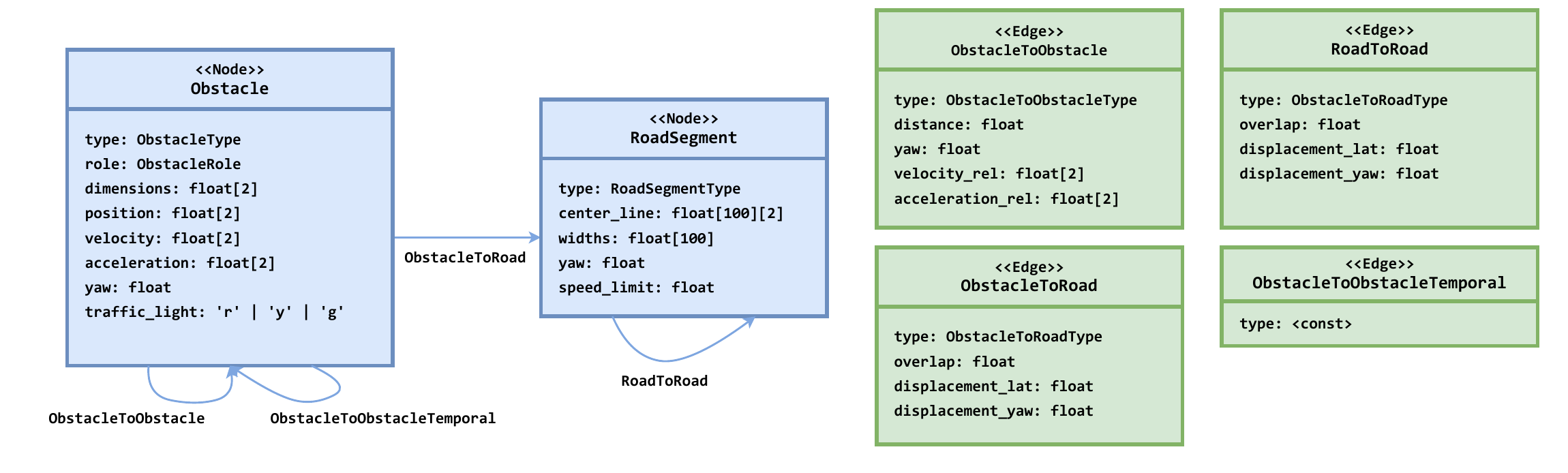}
    \centering
    \caption[Heterogeneous scenario graph data model]{Heterogeneous scenario graph data model}
    \label{fig:scenario_graph_data_model}
\end{figure*}

Strongly inspired by HDGT \cite{jia_hdgt_2023} and HoliGraph \cite{grimm_heterogeneous_2023}, we propose the graphical data model depicted in \Cref{fig:scenario_graph_data_model,fig:scenario_graph}. Our graph is heterogeneous in that it consists of map- and obstacle (static and dynamic) nodes (named \texttt{RoadSegment} and \texttt{Obstacle}) and four types of edges (subtypes listed in \Cref{tab:edge_types}) to encode relations among road segments and obstacles respectively as well temporal edges in the sense of temporal unrolling (see \Cref{sec:2.1}). Obstacle nodes feature the respective obstacles' states at a point in time, including, among others, its type, role (\texttt{static}, \texttt{dynamic}) and pose. Road segments are primarily defined by a fixed-length sequence of centerline points, accompanying widths and a type (\texttt{lanelet, walkway, other}). HDGT \cite{jia_hdgt_2023} found relative positions to add tremendous expressiveness over absolute ones. Consequently, we encode coordinates relative to a common reference point, which we chose as the median position of all obstacles through the whole scenario. This ensures translation invariance. In addition, edge attributes encode pair-wise relative positions and velocities to facilitate rotation invariance. Accordingly, our representation is scene-centric. Temporal / sequential information is only encoded implicitly through the presence or absence of temporal edges, as well as sinusoidal positional encoding added onto the node feature vector. To increase a node's receptive field, a configurable parameter (called \textit{temporal reach} with a default value of 4) permits temporal edges of multiple hops to the past. This is conceptually similar to skip connections, hence allowing to gather information beyond the scope of standard message passing. Road segments are only included within a range of the scenario's effective boundaries plus a buffer of 100 m. Also, they are entirely optional, allowing to represent ``off-road'' situations as well. A key characteristic of this representation is the ability to encode an entire, variable-length scenario in a single data structure and thus, later on, as a single vector. 

\begin{table}[]
    \caption{Edge types of proposed scenario graph}
    \centering
    \renewcommand{\arraystretch}{1.25}
    \resizebox{\columnwidth}{!}{%
    \begin{tabular}{@{}l p{0.6\columnwidth} @{}}
    \toprule
    \textbf{Edge}           & \textbf{Subtypes} \\ \midrule
    \textbf{\texttt{RoadToRoad}}         & \texttt{predecessor}, \texttt{successor}, \texttt{adj\_left}, \texttt{adj\_right}, \texttt{merging}, \texttt{diverging}, \texttt{intersecting}, \texttt{other} \\
    \textbf{\texttt{ObstacleToObstacle}} & \texttt{behind}, \texttt{in\_front}, \texttt{left}, \texttt{right}, \texttt{same\_lane}, \texttt{must\_yield\_row}, \texttt{must\_yield\_tl}, \texttt{other}   \\
    \textbf{\texttt{ObstacleToRoad}} & \texttt{is\_on}, \texttt{is\_close} \\
    \textbf{\texttt{ObstacleToObstacleTemporal}} & - \\ \bottomrule
    \end{tabular}%
    }
    \label{tab:edge_types}
\end{table}

\subsection{Similarity Learning Architecture}
\label{sec:3.2}
Following the common scheme in similarity learning, our goal is to obtain expressive vector representations of scenario graphs in an embedding space where vector distances mirror similarity. These embeddings can then be used for clustering using well-established algorithms. In the absence of labels, we are compelled conduct training in an unsupervised or self-supervised manner. Autoencoders and other generative models are a popular instrument in this regard, however, they necessitate a decoder component, which, especially in view of the graph isomorphism problem, is not trivial to design.

\subsubsection{Bootstrapping model (BGRL)}
\begin{figure}
    \includegraphics[width=90mm]{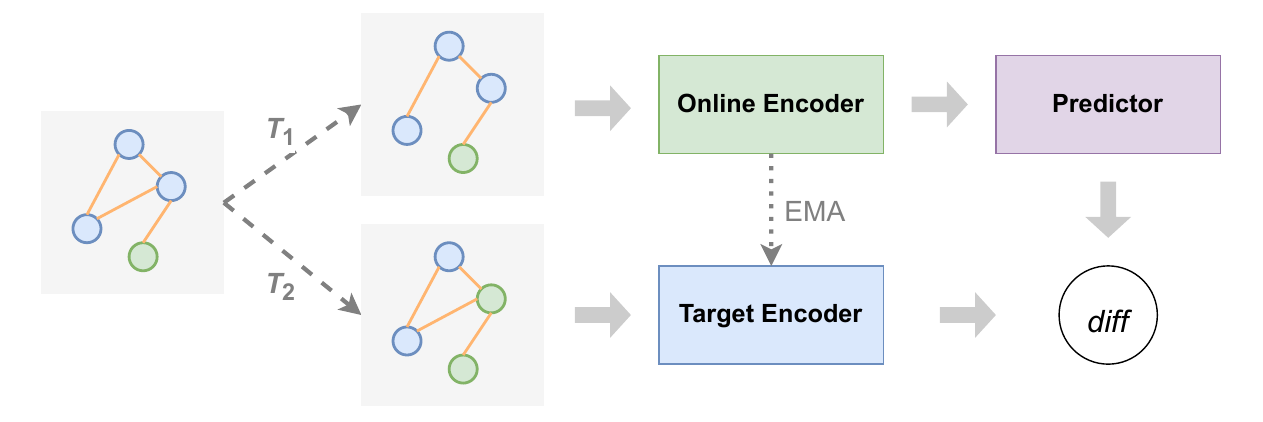}
    \centering
    \caption[BGRL architecture schema, inspired by \cite{thakoor_large-scale_2023}]{BGRL architecture schema, inspired by \cite{thakoor_large-scale_2023}}
    \label{fig:bgrl}
\end{figure}

Because we are only facing a problem of discriminative nature and thus have no actual necessity for a decoder, contrastive or bootstrapping-inspired methods are a more obvious choice. Accordingly, our proposed architecture follows that of BGRL \cite{thakoor_large-scale_2023}, depicted in \Cref{fig:bgrl}, but extends it in two aspects. In its original formulation, embeddings are produced per node and for homogeneous graphs (also without edge features), whereas we need graph-level representations of heterogeneous graphs instead. Therefore, we introduce additional pooling to the encoder and extend it to feature multiple parallel convolutions for heterogeneity. 

\begin{figure*}[t!]
    \includegraphics[height=62mm]{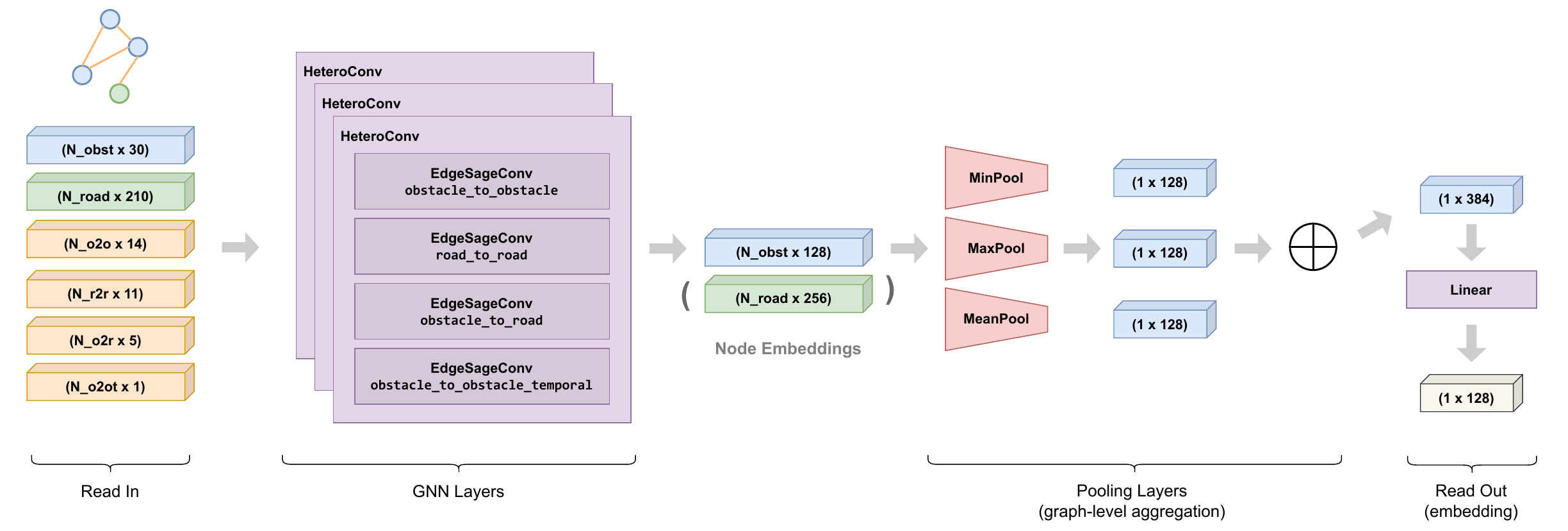}
    \centering
    \caption[Heterogeneous graph encoder architecture]{Heterogeneous graph encoder architecture}
    \label{fig:encoder_architecture}
\end{figure*}

Conceptually, the model works like this: two augmented versions of the same one graph are fed into an \textit{online encoder} (O) and an identically structured \textit{target encoder} (T) network respectively. The embedding produced by O is fed through another simple multi-layer perceptron named the \textit{predictor} (P), which is trained with the objective to predict the output of O. The loss is propagated back through P and O, while the parameters of T are frozen. Instead, T is updated from O at regular intervals using an exponential moving average (EMA). The authors show that O learns to map to a descriptive embedding space in terms of similarity and we observe this concept to remain still functioning on a graph-level as well.

At each step during training, we sample two different sets of graph-level augmentations. Inspired by \cite{thakoor_large-scale_2023}, they are chosen as a combination of probabilistic (1) edge dropping, (2) attribute dropping and (3) attribute perturbation (using standard Gaussian noise), parameterized with a binomial probability of \(0.1 \le p \le 0.2\).

Cosine similarity between the L2-normalized target- and predicted embeddings (and vice versa) is used as a symmetric loss. Deviating from the original paper, the target network weights are only updated every \textit{k}-th step (with the original "reverse"-decaying momentum), whereby we chose \(k = 10\).

\subsubsection{Contrastive model (GraphCL)}
In addition to the BGRL approach, we implement a second embedding model that employs contrastive learning based on the GraphCL \cite{you_graph_2020} architecture and conduct a brief comparison between the two. The model's architecture consists of a (GNN-) encoder network, followed by a linear projector and operates on two augmented variants of the same graph (positive pair) and another augmented negative sample respectively. As commonly done, the negative candidate is simply chosen as an arbitrary other sample from the same batch. The same set of augmentations and an identical encoder as with the BGRL variant are reused (see \Cref{sec:encoder} below). Following the original implementation, the loss is computed with the objective to minimize distance between embeddings of the positive pairs while maximizing the distance between a pair of positive and negative sample. 

\subsubsection{GNN encoder}
\label{sec:encoder}
Our GNN encoder, depicted in \Cref{fig:encoder_architecture}, consists of three convolution layers for each of the four edge types with batch normalization and ReLU activations in between. For obstacle embeddings, hidden dimensions are chosen as 32, 64 and 128, for road segment nodes we use 64, 128 and 256 respectively. Graph convolutions are followed by three parallel pooling layers (min, max and mean) (inspired by \cite{lee_generalizing_2016}), that squash the embeddings of \texttt{Obstacle} nodes to three fixed-size vectors. These are concatenated horizontally and fed through a final linear layer to produce the 128-dimensional graph-level output. The predictor P is a simple 2-layer feed-forward network with a hidden dimension of 512 and PReLU activation in between. We employ a customized version of PyTorch-Geometric's\footnote{\url{https://pytorch-geometric.readthedocs.io}} \texttt{SAGEConv} operator that additionally supports to incorporate edge features in message passing:

\vspace{-0.2cm}
\begin{equation}
    x_i' = W_1 x_i + W_2 \cdot \frac{1}{|\mathcal{N}(i)|} \sum_{j \in \mathcal{N}(i)} \left( x_j + W_e e_{ij} \right)
    \label{eq:edgesageconv}
\end{equation}

All hyperparamteres were determined empirically, details on the analyses are omitted for brevity.

Eventually, density-based clustering with HDBSCAN is employed to discover semantically similar groups of scenarios, given their embedding vectors.

\section{Evaluation}

\subsection{Experiment Settings}
As for the evaluation of our previously introduced concept, we particularly seek to investigate two aspects. First, we aim to quantify the expressiveness of the learned embedding space with respect to scenario similarity. To this end, we first examine the degree to which the overall training objective is satisfied, that is, the percentage of test samples for which a graph and its augmented version are located closer in the embedding space than compared to an arbitrarily chosen other graph (see \Cref{eq:embedding_validity}). Although this is a rather trivial criterion, it is nevertheless a necessary prerequisite for the entire approach to succeed. 

\vspace{-0.3cm}
\begin{equation}
    d(G_1, \tau(G_1)) < d(G_1, G_2)
\label{eq:embedding_validity}
\end{equation}

Moreover, inspired by Thakoor et al. \cite{thakoor_large-scale_2023}, we evaluate with respect to a downstream supervised learning task. In particular, we train a simple multi-class, multi-label classifier to predict a scenario's set of labels as obtained from the nuPlan dataset\footnote{\url{https://www.nuscenes.org/nuplan}} and feed solely the respective embedding vectors as an input. Intuitively, if the classifier is able to extract meaningful information from the embeddings, we get a good indication about their quality. We reuse the predictor network's simple architecture from \Cref{sec:3.2} for the classifier and train it with a binary cross entropy loss on the output logits. Besides the designated test split, we additionally conduct this evaluation on a holdout set comprising previously unseen locations to gain insights about generalizability. Primarily, we measure multi-label accuracy, whereby the set of predicted labels must not be an \textit{exact} match, but rather a superset of the ground-truth labels. Moreover, we measure AUPRC (area under precision-recall curve) averaged across all samples to gauge the model's capability to correctly identify a high percentage of true positives while also rejecting enough false samples. 
However, it is worth mentioning that the nuPlan labels only serve as a rough proxy. Besides being very fine-grained and partially ambiguous, they are constructed from an ego-centric perspective, whereas our scenarios are of scene-centric nature instead. Hence, the above metrics are representative only to a certain extent.

The second part of the evaluation concerns clustering. To gain insights about the clustering results' quality in general, we compute silhouette coefficient as a generic metric. In addition, we consult the previously mentioned class labels again to gauge the correspondence between clusters and ground-truth classes. We adopt the notion of ``overall accuracy'' from Weber et al. \cite{weber_toward_2023} and slightly modify it for our multi-label problem, see \Cref{eq:multilabel_acc}. For each cluster, we determine the ground-truth label that is most prominent and declare it as the cluster label. We then count the percentage of all samples whose label set contains a label that matches the label of the cluster they were assigned to. Note that we do not require an exact match of \textit{all} labels of a sample but instead settle with a ``at-least-one'' logic. Both metrics (silhouette score and accuracy) are only computed on the subset of data points that are explicitly assigned to a specific cluster, disregarding unclustered samples.

\begin{figure}
\begin{equation}
    \textit{multilabel\_acc} = \frac{1}{N} \sum_{i=1}^{N} \mathbbm{1} \left( y_i \cap \hat{y}_{c(i)} \neq \emptyset \right)
    \label{eq:multilabel_acc}
\end{equation}
\vspace{-0.2cm}
\caption*{\small \Cref{eq:multilabel_acc}: ``Multilabel'' accuracy used to evaluate clustering, where \(c(i)\) is a sample's assigned cluster, \(y_i\) is a sample's ground-truth label set, \(\hat{y}_{j}\) is the primary label of cluster \(j\) (single-element set), and \(\mathbbm{1}\) is the binary indicator function.\normalsize}
\end{figure}

\subsection{Dataset}
All experiments use 25,000 scenarios sampled from the nuPlan dataset's \texttt{boston} split in a \(85:15\) train-test-split. Downstream classification performance is additionally cross-validated on 3,750 scenarios from the \texttt{pittsburgh} and \texttt{singapore} splits respectively. We chose a subset of ten labels among the most common nuPlan labels (e.g. \small\texttt{near\_pedestrian\_on\_crosswalk}, \texttt{high\_magnitude\_speed}, \texttt{starting\_left\_turn}\normalsize). Since every scenario can have \( [1..10] \) labels, there are \(2^{10} = 1024\) possible combinations, however, only 69 distinct label sets are actually contained in our subset of data. 

\subsection{Embedding Space}
After training the BGRL- and GraphCL-inspired models introduced in \Cref{sec:3.2} for 50 epochs each with a batch size of 32 using the AdamW optimizer with a learning rate and weight decay of 0.001, we observe \text{99.99 \%} compliance with \Cref{eq:embedding_validity} for both models and achieve the classification results listed in \Cref{tab:classification}. Specifically, an accuracy of 59.9 \% and 44.7 \% was achieved for the bootstrapping- and contrastive methods on the \texttt{boston} split respectively. Notably, despite a perceptible drop in accuracy, the downstream model is nonetheless able to generalize to previously unseen environments. Although both models perform decently well, BGRL generally appears to perform better, i.e. construct richer embeddings across all dataset splits on average. 

\begin{table}[]
    \caption{Downstream multi-label classification performance}
    \centering
    \small
    \renewcommand{\arraystretch}{1.2}
    \begin{tabular}{@{}llll@{}}
    \toprule
    \textbf{Dataset} & \multicolumn{1}{l}{\textbf{Model}} & \multicolumn{1}{l}{\makecell[l]{\textbf{Accuracy} \\ \footnotesize(``contain'')\normalsize}} & \multicolumn{1}{l}{\textbf{AUPRC}} \\ \midrule
    \multirow{2}{*}{\footnotesize\texttt{boston}\normalsize}     & BGRL    & 0.599 & 0.757 \\
                                         & GraphCL & 0.447 & 0.666 \\ \midrule
    \multirow{2}{*}{\footnotesize\texttt{pittsburgh}\normalsize} & BGRL    & 0.477 & 0.563 \\
                                         & GraphCL & 0.412 & 0.478 \\ \midrule
    \multirow{2}{*}{\footnotesize\texttt{singapore}\normalsize}  & BGRL    & 0.341 & 0.488 \\
                                         & GraphCL & 0.289 & 0.399 \\ \bottomrule
    \end{tabular}
    \label{tab:classification}
\end{table}

Given these results, we conclude that not only the training objective was met, but the learned latent space also appears to be sufficiently expressive to carry useful properties about scenarios. Note that the absolute accuracy scores are not of actual relevance, but rather is the fact that a classifier model can, in fact, be trained in the first place.

\subsection{Clustering}

\captionsetup{justification=centering}
\begin{figure}
    \hspace*{-0.3cm}
    \includegraphics[width=90mm]{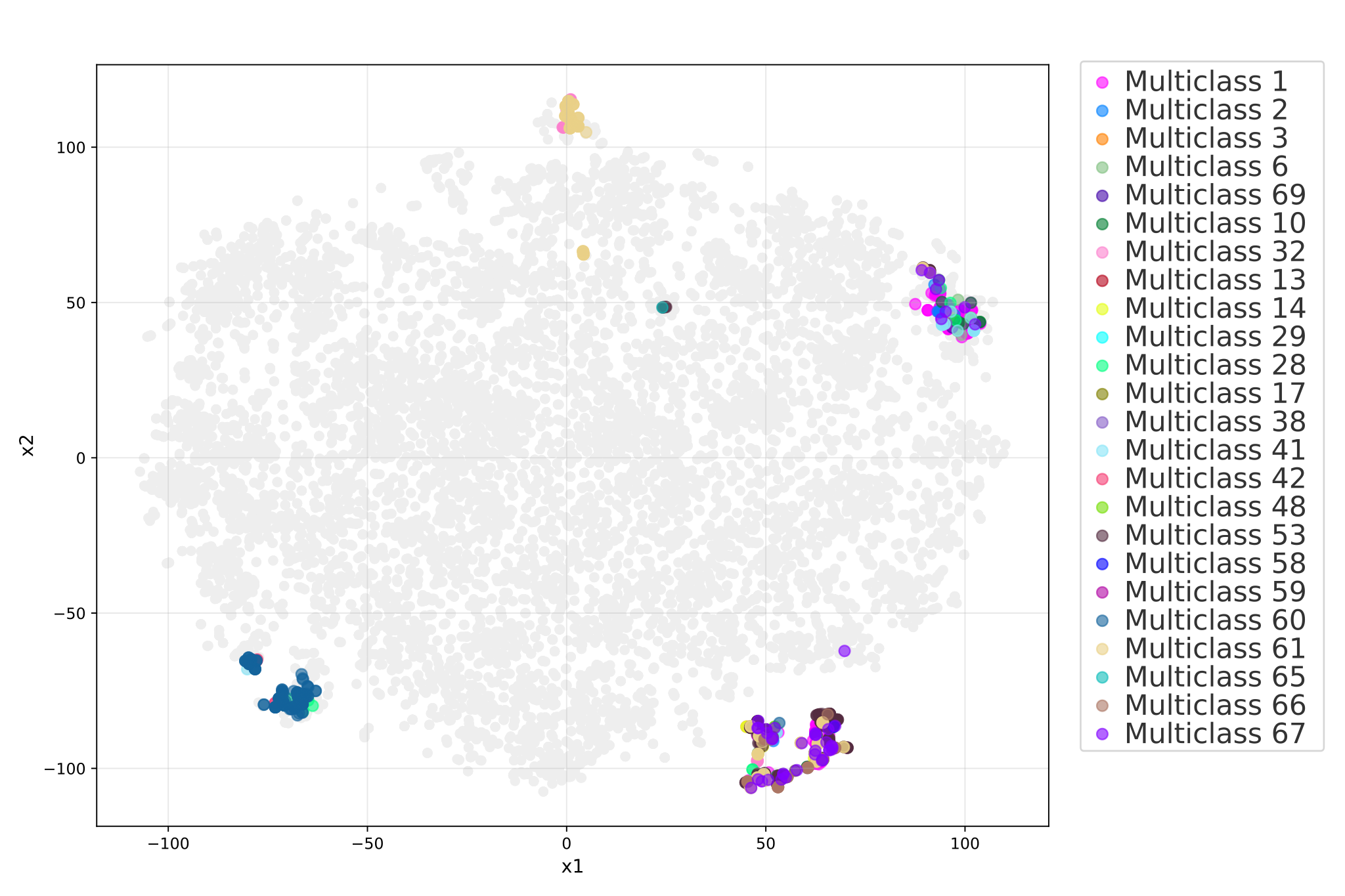}
    \vspace{-0.6cm}
    \caption{Subset (seven randomly selected clusters) of the clustered embeddings space with colors corresponding to original ground-truth class label sets}
    \label{fig:clustering_space}
    \vspace{-0.2cm}
\end{figure}

\begin{figure}
    \includegraphics[width=86mm]{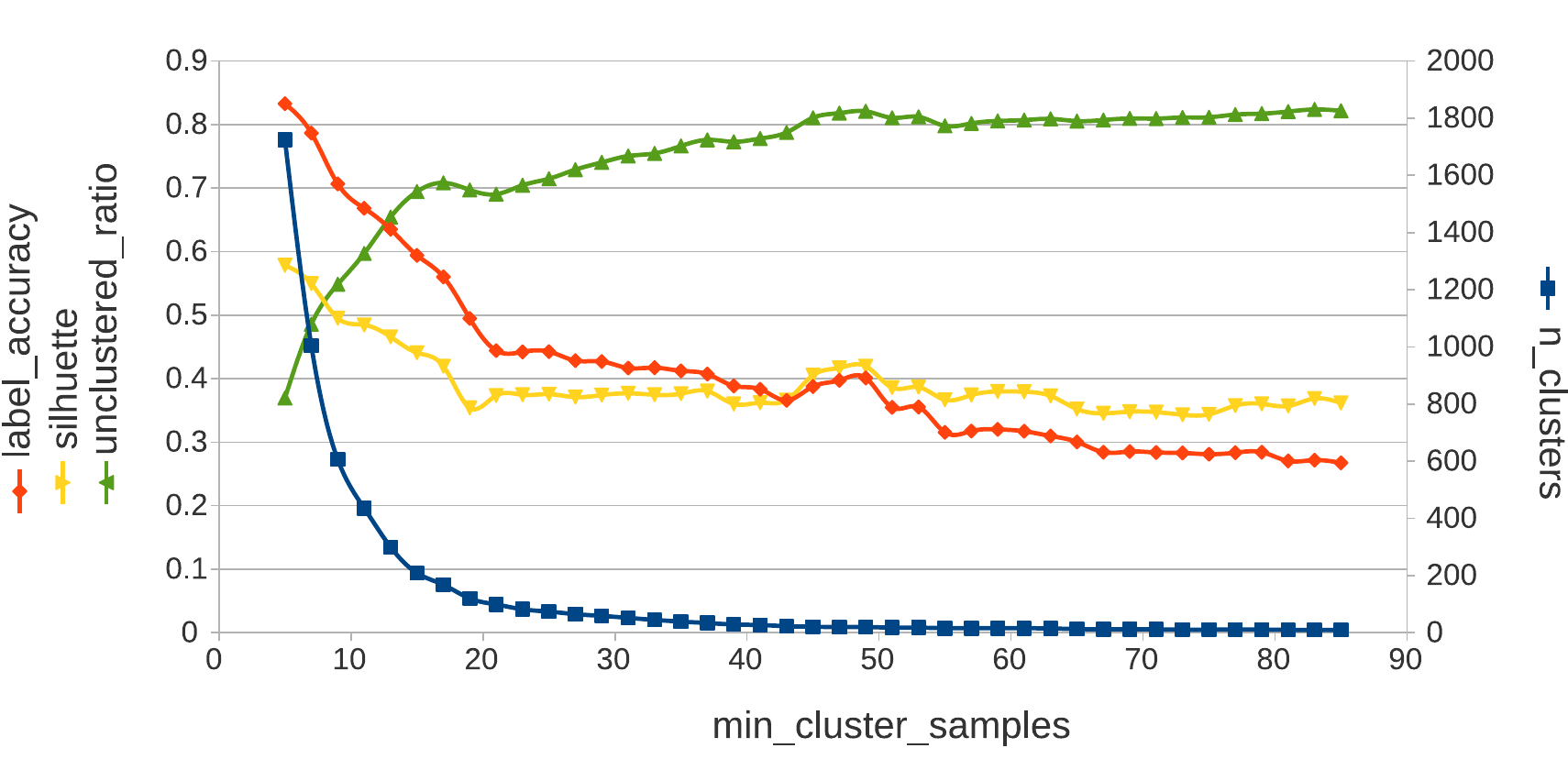}
    \vspace{-0.2cm}
    \caption{Hyperparameter search for \texttt{min\_cluster\_size}. Percentage of unassigned points increases for increasingly large minimum cluster size.}
    \label{fig:clustering_bgrl}
    \vspace{-0.3cm}
\end{figure}

The most influential hyperparameter of the HDBSCAN clustering algorithm is \texttt{min\_cluster\_size} (\texttt{mcs} for short), that is, the minimum number of samples in a group for that group to be considered a cluster. This parameter significantly affects the total number of clusters found and the percentage of points that remain unassigned. There is no objective right or wrong choice, but instead, it must be set in accordance with the concrete use case at hand. A hyperparameter search yields the results shown in \Cref{fig:clustering_bgrl}. The highest silhouette score (0.57) and accuracy (0.83) and the lowest ratio of unclustered points (36.9 \%) are obtained for \text{\texttt{mcs = 5}}, however, with an unmanageably large number of 1724 clusters, which can not be made sense of in a meaningful way. Guided by the number of 69 ground-truth classes, we instead choose a setting that leads to an approximately similar number of clusters at \text{\texttt{mcs = 25}} and obtain a multi-label accuracy of 0.44 and a silhouette score of 0.38. In comparison, the embeddings produced by our contrastive model give similar accuracy but yield more cohesive, denser clusters, as outlined in \Cref{tab:clustering_results}. Given this aforementioned parameterization, we obtain a decently large set of groups that appears realistic to be interpreted as distinguishable scenario classes, even though the resulting metrics still indicate room for improvement in clustering quality. A noticeable characteristic of the clustering space (\Cref{fig:clustering_space}) is the exceedingly large portion of unclustered points. In our interpretation, these scenarios correspond to just ``normal'' driving that is not sufficiently distinguishable. While some scenarios exhibit striking features, the majority of them is rather a ``blend'' of multiple types.

\begin{table}[]
    \caption{Embeddings clustering results}
    \centering
    \small
    \renewcommand{\arraystretch}{1.2}
    \begin{tabular}{@{}llllll@{}}
    \toprule
    \textbf{\texttt{mcs}} & \textbf{Model} & \textbf{\# Clusters} & \textbf{Unclust.} & \textbf{Acc.} & \textbf{Silhouette} \\ \midrule
    \multirow{2}{*}{\textit{\textbf{5}}}  & BGRL    & 1724 & 36.9 \% & 0.833 & 0.578 \\
                                          & GraphCL    & 1346 & 44 \%   & 0.813 & 0.53  \\ \midrule
    \multirow{2}{*}{\textit{\textbf{25}}} & BGRL & 74   & 71.4 \% & 0.442 & 0.375 \\
                                          & GraphCL & 68   & 58.8 \% & 0.443 & 0.584 \\ \bottomrule
    \end{tabular}
    \label{tab:clustering_results}
\end{table}

\subsection{Qualitative results}
\begin{center}
\begin{figure}
    \includegraphics[width=88mm]{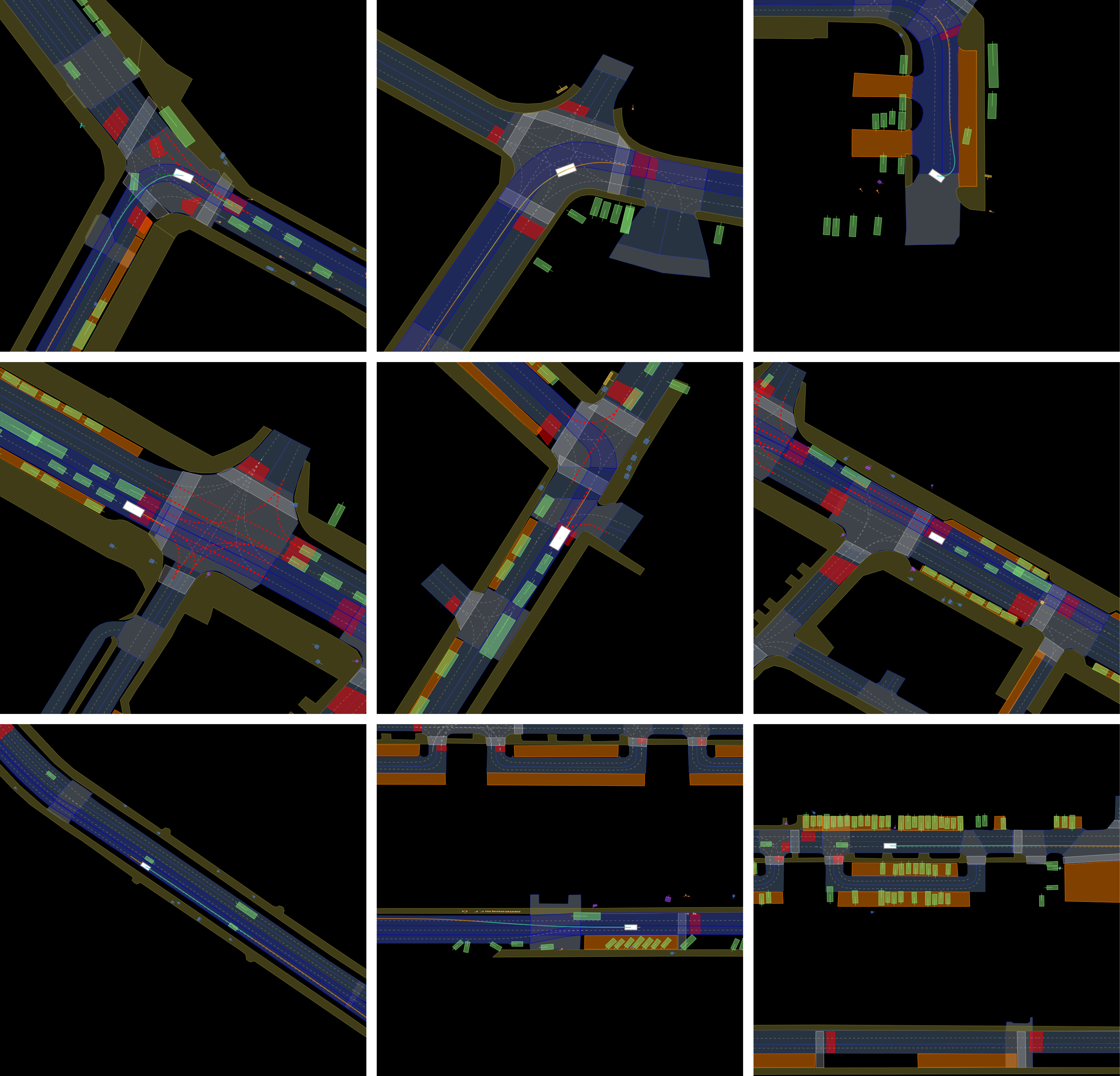}
    \caption[]{Exemplary scenarios drawn from three different clusters (ego in white) \\ \vspace{0.1cm} \small\textbf{1st row:} multiclass 65 (\texttt{starting\_left\_turn}) \\ \text{\textbf{2nd row:} multiclass 61 (\texttt{on\_stopline\_traffic\_light})} \\ \textbf{3rd row:} multiclass 11 (\texttt{high\_magnitude\_speed}) \normalsize}
    \label{fig:scenarios_examples}
    \vspace{-0.5cm}
\end{figure}
\end{center}

Given the ultimate objective of organizing semantically similar scenarios into distinct classes, the interpretability of these classes is of particular interest. Therefore, we visually inspect samples drawn from different clusters to find commonalities and differences between them. Indeed, we found many clusters to correspond to clearly distinguishable maneuvers. For instance, the first row of scenario snapshots in \Cref{fig:scenarios_examples} depicts situations in all of which the ego performs a left turn. This matches the fact that the respective cluster's most prominent label is, in fact, \texttt{starting\_left\_turn}. In the second group, all scenarios portray the ego stopping at a traffic light, while in the third group, the ego drives on a long straight road. Remarkably, the embedding space appears to have been learned in a rotation-invariant way, since similar scenarios are grouped regardless of the scene's orientation. On the other hand, however, we also observe clusters, that do not unveil any obviously interpretable coherence between their scenarios.

\vspace{-0.5cm}
\section{Conclusion \& Outlook}

This work proposes a method to facilitate large-scale scenario-based testing by structuring the otherwise impractically large search space of test cases into distinct, representative groups of scenarios. First, complex, variable-length scenarios are encoded using the heterogeneouos, spatio-temporal graph model proposed in \Cref{sec:3.1}. Second, a latent-space vector representation is learned that allows for similarity search among these scenario graphs. This is achieved in a self-supervised manner without incorporating expert knowledge or other labels and without limiting oneself to some pre-defined criterion or ODD. Specifically, we adapt the bootstrapping architecture to a graph level, equip the model with a capable encoder to cope with heterogeneity and apply it as a similarity-aware feature extractor. Quantitative analyses confirm the fact that embeddings carry meaningful information about scenarios in order to distinguish between different types of such. Lastly, we demonstrate how to effectively partition the scenario space by applying clustering to obtain distinct groups. Qualitative and quantitative observations affirm these clusters to correspond to distinguishable driving maneuvers and situations. Nevertheless, clustering performance leaves significant room for improvement to become profitable for real-world use cases.

Based on our findings, future research may conduct more in-depth analyses on the embedding space and further refine it to improve clustering performance. Another desirable property of the embedding space to aim for is to allow for smooth interpolation between clusters to obtain ``blends'' of multiple discrete scenario types. With regard to evaluation, it would be interesting to investigate performance on a greater variety of datasets and, inspired by Zipfl et al. \cite{zipfl_traffic_2023}, gain better interpretability of the embeddings by investigating their correlation with certain scenario attributes or criticality. Moreover, inspired by ScenarioNet \cite{li_scenarionet_2023}, running experiments in conjunction with an end-to-end AD stack like openPilot\footnote{\url{https://github.com/commaai/openpilot}} would be desirable. Ultimately, the overarching goal is to use the scenario embeddings as guidance / conditioning in (graph-) generative to synthesize entirely new scenarios for simulation. 

\section*{Acknowledgment}
The work leading to these results was supported under the ``DigiT4TAF'' service contract, funded by the German Ministry of Transport, Baden-Württemberg.

\bibliographystyle{IEEEtran}
\bibliography{references}

\end{document}